\documentclass[letterpaper, 10 pt, conference]{ieeeconf}
\IEEEoverridecommandlockouts
\usepackage{cite}
\usepackage{amsmath,amssymb,amsfonts}
\usepackage{bm}
\usepackage{algorithmic}
\usepackage{graphicx}
\usepackage{textcomp}
\usepackage{xcolor}
\usepackage{subfig}
\usepackage{todonotes}
\usepackage{hyperref}

\def\BibTeX{{\rm B\kern-.05em{\sc i\kern-.025em b}\kern-.08em
    T\kern-.1667em\lower.7ex\hbox{E}\kern-.125emX}}
    
\newcommand{\uvec}[1]{\boldsymbol{\hat{\textbf{#1}}}}

\begin{document}

\title{\LARGE \bf
A Reactive Autonomous Camera System for the \\ RAVEN II Surgical Robot\\
}


\author{Kay Hutchinson$^1$, Mohammad Samin Yasar$^1$, Harshneet Bhatia$^2$, Homa Alemzadeh$^1$
\\$^1$Department of Electrical and Computer Engineering, University of Virginia, Charlottesville, VA  22904 \\
$^2$Department of Computer Science, University of Virginia, Charlottesville, VA 22904 \\
\{kch4fk, msy9an, hb9ee, alemzadeh\}@virginia.edu
}

\maketitle

\begin{abstract}
The endoscopic camera of a surgical robot provides surgeons with a magnified 3D view of the surgical field, but repositioning it increases mental workload and operation time. Poor camera placement contributes to safety-critical events when surgical tools move out of the view of the camera. This paper presents a proof of concept of an autonomous camera system for the Raven II surgical robot that aims to reduce surgeon workload and improve safety by providing an optimal view of the workspace showing all objects of interest. This system uses transfer learning to localize and classify objects of interest within the view of a stereoscopic camera. The positions and centroid of the objects are estimated and a set of control rules determines the movement of the camera towards a more desired view. Our perception module had an accuracy of 61.21\% overall for identifying objects of interest and was able to localize both graspers and multiple blocks in the environment. Comparison of the commands proposed by our system with the desired commands from a survey of 13 participants indicates that the autonomous camera system proposes appropriate movements for the tilt and pan of the camera.
\end{abstract}


\section{Introduction}
Surgical robots such as Intuitive Surgical's da Vinci Systems \cite{daVinci} are advancing medical specialties such as urology, gynecology, and general surgery by providing surgeons with increased flexibility and precision, while reducing incision size, recovery time, and scarring. Their adoption is linked to an increase in volume of minimally invasive surgery (MIS) cases \cite{barbash2010new}, and is driving the development of new surgical procedures and technologies \cite{peters2018review}.
However, the current generation of robots is not autonomous yet. They are in level 0 of autonomy \cite{yang2017medical} and the surgeon must position and control all four arms manually which increases their mental workload \cite{pandya2014review}. One of these arms holds the endoscopic camera which provides surgeons with a magnified 3D view of the surgical field, but requires both hands and a foot to switch control of the arms and reposition the camera. During an operation, surgeons often adjust their camera position or settle for a suboptimal viewpoint which increases procedure time and risk to the patient since poor camera placement contributes to safety-critical events such as arm collisions, use of excessive force, dropping an object, or movement of the instruments out of camera's view \cite{alemzadeh2014systems, alemzadeh2016adverse}. 

Existing methods for automating the surgical robot's camera focus on using \emph{reactive} simple sets of rules, \emph{proactive} machine learning algorithms to learn movement behavior, or \emph{combined control} strategies that integrate these two techniques \cite{pandya2014review}. Most reactive autonomous camera systems rely on kinematic data from the surgical robot \cite{mudunuri2010autonomous}, use a camera to track the surgical tools \cite{omote1999self}, or track the surgeon's eyes using visual servoing \cite{wei1997real}, and change the camera position in direct relation to changes in these measurements. But, these methods over-emphasize tracking tools and do not account for other objects of interest in the environment such as important tissues or needles. Our goal is to create an autonomous camera system that reduces surgeon workload and improves safety by providing an optimal viewpoint of the surgical environment that keeps \emph{objects of interest and surgical tools} in the field of view thus reducing the likelihood of off-camera injuries.

In related work on autonomous camera systems, Yu et al. \cite{yu2016automatic} introduced a region of interest (ROI) around the robot end effectors. 
Yang et al. \cite{yang2019adaptive} defined an intuitive virtual plane (IVP) as the plane normal to the surgeon's line of sight and containing the intersection of the surgeon's line of sight with the ROI. The IVP was a constraint to reduce misorientation that occurs when the optical and physical axes of the laparoscope are not parallel. However, their work used a 2D laparoscope and followed only one end effector.

While previous work focused on laparoscopic surgery,~\cite{king2013towards, eslamianautonomous, eslamian2019development} were the first works that created an autonomous camera system for a surgical robot using the da Vinci Research Kit (DVRK) \cite{kazanzides2014open}. Their system used several rules and kinematics data from the DVRK to keep the tools centered in the camera's view. They conducted a 20-participant trial, which included four surgeons, that compared their automated camera system to the traditional clutched camera control. The results showed that the automated camera was able to keep the tools within the camera's field of view, and improved metrics of workload, efficiency, and progress. In contrast, our work relies on video data for detecting objects in the surgical workspace, thus enabling the consideration of \emph{all} objects of interest in the environment as well as being platform independent.

\textbf{Contributions.} 
This paper presents a proof of concept of an autonomous camera system for the RAVEN II robot, an open-source platform for robotic surgery research \cite{hannaford2012raven}.
\begin{itemize}
    \item We introduce a custom-built camera arm, the Independent Binocular Imaging System (IBIS) (Section \ref{subsubection: IBIS}), supporting a 3D stereoscopic camera (ZED Mini by Stereolabs Inc. \cite{zed}). The IBIS reports its joint and camera positions, and accepts commands from foot pedals and serial communication which facilitates system integration.
    \item We present a Perception module for automated perception of the surgical field using transfer learning to localize and classify objects of interest (end effectors and blocks in dry-lab) in a given image of the surgical field. We use a Mask Region-based CNN (MRCNN) \cite{he2017mask} (with convolutional layers pre-trained using the COCO dataset \cite{matterport_maskrcnn_2017}) to identify objects of interest by generating bounding boxes, classifying, and then creating masks. The coordinates of all objects of interest provided by the Perception module are then used to calculate the centroid of the objects and to adjust the camera position (Section \ref{subsection: Position Estimation}).
    \item We present a Control module that adjusts the zoom, pan, and tilt of the camera to align the center of the view with the centroid of the objects and maximize the view of all objects of interest. The Control module computes the projection of each object's position onto an image plane containing the centroid. Then, control rules determine camera movements based on the position of the centroid relative to the desired view area of the camera.
\end{itemize}

The training set for the MRCNN consisted of 1,686 annotated images (3,372 after augmentation), and the validation set consisted of 600 images. The images were collected from dry-lab experiments of the ``Pick and Place" task. These datasets included both left and right images from the ZED Mini camera, but there was not a 50/50 ratio between the two. 
After tuning the hyper-parameters using the early stopping technique, the overall loss (defined as the sum of the region proposal class loss, region proposal bounding box loss, MRCNN class loss, MRCNN bounding box loss, and MRCNN mask loss) was 0.2672. 

We evaluated the camera system using a separate set of 27 pairs of left and right images of a block transfer task, and a survey was used to determine the ground truth desired commands for each image. 
The Perception module correctly identified 61.21\% of the objects, and the Control module demonstrated acceptable behavior when tilting and panning, but the desired zoom area should be decreased to provide a wider field of view.



\section{Autonomous Camera System}
\begin{figure}[htbp]
\centerline{\includegraphics[width=0.45\textwidth]{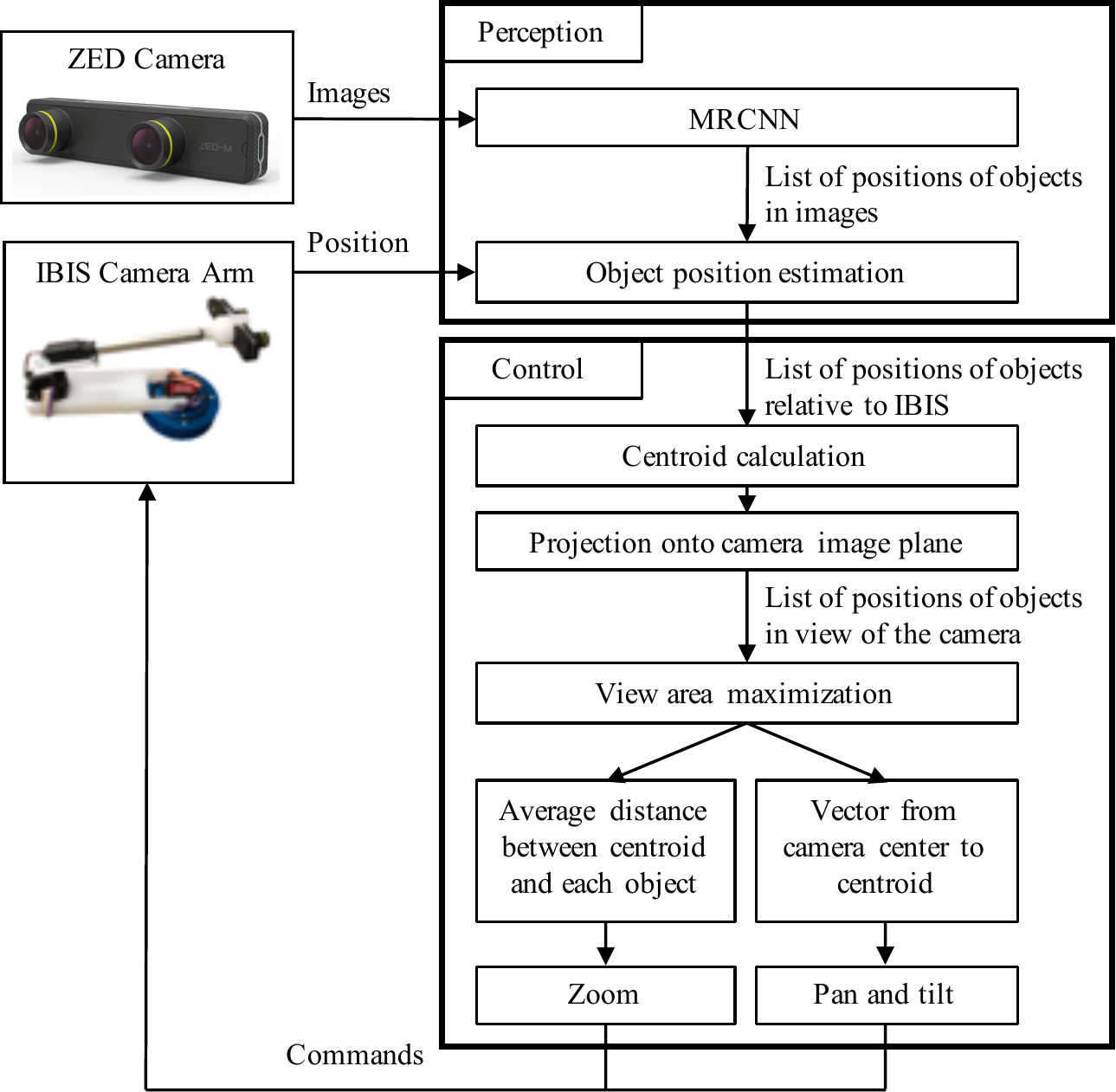}}
\caption{Autonomous camera pipeline}
\label{pipeline}
\vspace{-2em}
\end{figure}

Our autonomous camera system consists of a custom-built camera arm integrated with a ZED Mini and a software pipeline for perception of the environment and control of the camera arm, as shown in Fig.~\ref{pipeline} and described next. 

\subsection{IBIS Camera Arm}
\label{subsubection: IBIS}
\begin{figure}[htbp]
\vspace{-1.5em}
\centerline{\includegraphics[width=0.35\textwidth]{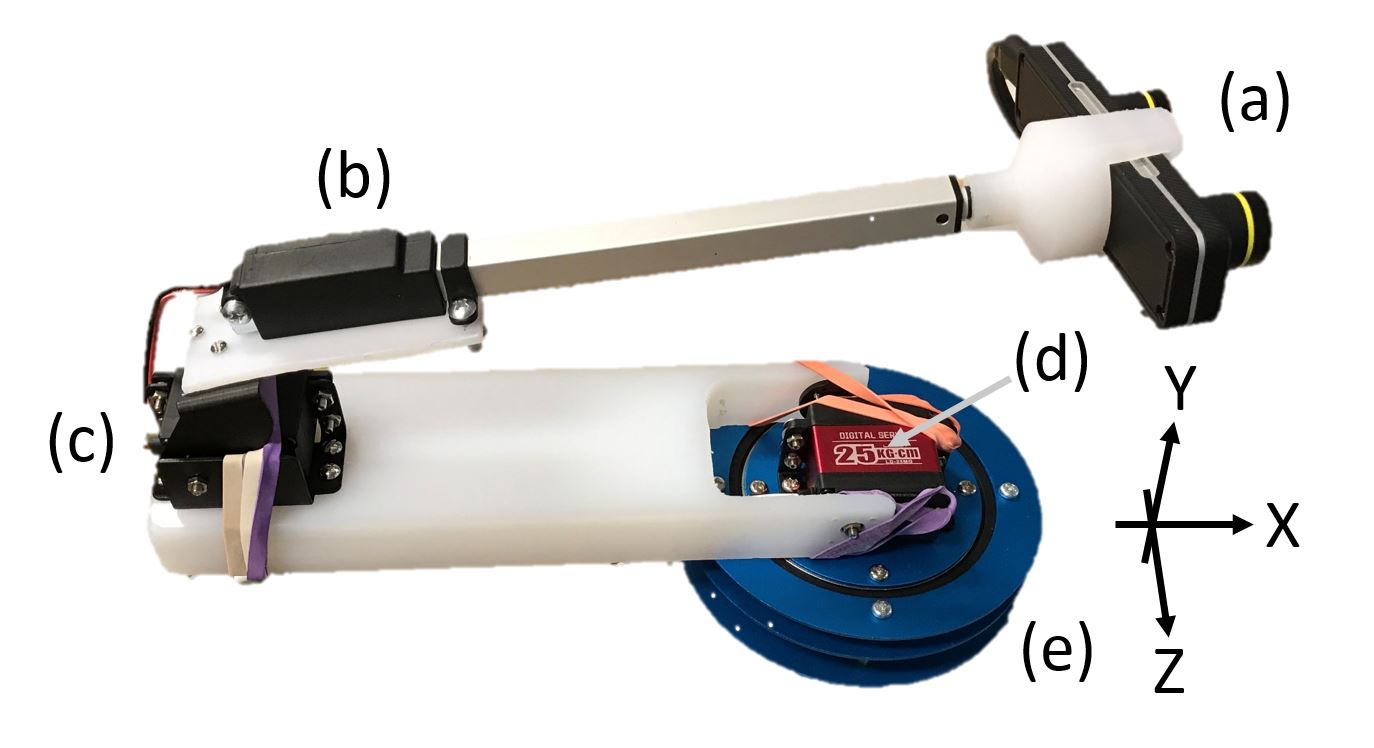}}
\vspace{-1em}
\caption{IBIS Camera Arm: (a) ZED Mini, (b) Linear Actuator, (c) Top Servo, (d) Bottom Servo, (e) Base and Base Servo.} 
\vspace{-0.5em}
\label{IBIS}
\end{figure}

A custom robotic camera arm, shown in Fig.~\ref{IBIS}, was developed to hold the stereoscopic camera and provide control over the position of the camera during tele-operation of the Raven II. This Independent Binocular Imaging System (IBIS) can also be controlled using a set of foot pedals making it platform independent and enabling future experimentation with alternative control interfaces for surgical endoscopes. An Arduino Uno R3 runs custom kinematic and inverse kinematic models, and communicates through a serial port to report its position and accept commands. This allows for open-source development and easy integration with other systems. 

The mechanical design of the IBIS consists of upper and lower arms positioned using three servos and a linear actuator. The Base Servo, embedded in the blue base of the IBIS shown in Fig.~\ref{IBIS}, controls the pan of the camera by rotating the entire arm left or right. The Bottom Servo is a high torque servo that supports the rest of the arm and camera. 
The Top Servo, at the joint between the upper and lower arms, is a standard servo with an angled bracket that joins to the mounting plate for the linear actuator. The ZED Mini is held by a custom-machined adapter that fits on the end of the linear actuator.

The Base Servo pans the camera arm, while the other two servos and linear actuator move together to change the vertical and horizontal position of the camera. These movements are calculated using the inverse kinematics functions based on incremental changes in the position of the Bottom Servo. This enables a deterministic solution for the positions of the Top Servo and linear actuator. The Base Servo is the origin of the coordinate system of the IBIS where the positive $x$ axis points into the operating space, the positive $y$ axis points upwards, and the positive $z$ axis points right.

The software on the Arduino Uno uses an open source library \cite{arduinoservo} to control the servos and the linear actuator. The main function of the code is to listen for commands from the foot pedals or serial line, use inverse kinematic functions to move the arm, update the state of the arm using the kinematic functions, and report the position of the arm over a serial line at 9600 baud rate. A Python script on the receiving computer parses and logs the position of the IBIS enabling real-time and closed loop control of the arm.

\subsection{Perception}
\label{subsection: Perception}
\subsubsection{Object Detection and Localization}
\label{subsection: Perception Detection}

We used Transfer Learning to apply the knowledge obtained from a pre-trained model to our task. This allowed us to obtain competitive accuracy with a smaller training dataset. As the perception task involves both object detection and localization, we used Mask Region-based Convolutional Neural Network (MRCNN) architecture, which has the backbone network, in our case pre-trained Residual Nets (ResNet) \cite{he2016deep}, followed by the network head. The backbone network performs feature extraction on a given image, and is followed by the Region Proposal Network (RPN) which examines each region of interest (ROI), before feeding the extracted features into the fine-tuned classification layers. The classification layers generate the bounding boxes and masks for each class. The bounding boxes indicate where an object has a high probability of being found while the masks reveal where the object was actually found. 


In our task, the classes of the objects were ``Left Grasper", ``Right Grasper", ``Red Block", ``Green Block", and ``Background". We fine-tuned the final two layers of the MRCNN, pre-trained on the COCO dataset \cite{matterport_maskrcnn_2017}, for our application of detecting and localizing objects of interest in the surgical workspace. 
The classification layers were trained on a set of 1,686 images (3,372 after augmentation) that were manually annotated using the VGG Image Annotator (VIA) \cite{dutta2016via}. The images were annotated by 5 students and each image was annotated once. Each object's boundary was outlined and it was labeled with the appropriate class. 
These images were collected using the ZED Mini from the execution of the Fundamentals of Laparoscopy (FLS) ``Pick and Place" task on the RAVEN II robot. 
The diversity in scenes was ensured to a certain degree by picking up different blocks with the graspers and showing incremental movements in the images. 

We performed image augmentation to diversify object orientations in the images of the training set and improve the model's ability to detect objects of interest.
After image augmentation, the training set consisted of 3,372 images. This is one method of artificially expanding a dataset. Some of the techniques that can be used for image augmentation include scaling, translation, rotation, flipping, adding noise, and changing lighting conditions. The techniques used for augmenting our dataset were flipping the images left/right 50\% of the time and generating images with random blending between the original images and their canny edges.

\subsubsection{Object Position Estimation}
\label{subsection: Position Estimation}
For each left and right image obtained from the ZED Mini, the MRCNN returns a list of the objects and the coordinates for the upper left and lower right corners of their bounding boxes. The centers of the bounding boxes are assumed to be the center of the object in the image, and are used for subsequent calculations. These objects are sorted and paired, and if an object was detected in one image but not the other, it is discarded because there is not enough information to reconstruct that object's position.

The locations of the objects in the left and right images are used to estimate each object's position with respect to the camera arm. Then, the centroid of the identified objects is calculated so that an image plane containing this point and perpendicular to the optical axis of the camera can be constructed. Each object is then projected onto this image plane in order to relate their locations to a desired field of view of the camera.

For each object in the list returned by the MRCNN, the center of the bounding box is calculated by taking the average of the $x$ and $y$ pixel coordinates of the corners. This estimates the center of each object in the left and right images. The lists are sorted to pair an object's location in the left image with its corresponding location in the right image. 

The position of the camera and its optical axis are obtained from the IBIS. A unit vector parallel to the camera's optical axis is defined as the normal unit vector, \(\uvec{n}\). The camera is assumed to be horizontal which allows the construction of a horizontal unit vector, \(\uvec{h}\), perpendicular to the optical axis of the camera. A third, orthogonal unit vector is defined as the cross product of the normal unit vector and the horizontal unit vector. This vector is named the vertical unit vector, \(\uvec{v}\). 

The distance, \(d\), of an object from the camera is calculated using the difference in the horizontal position of the object between the left and right images as shown in equation \eqref{objdistance}. The ZED Mini has a focal distance of \(f = 700\) pixels and a camera separation of \(S = 63\) mm. \(L_{px}\) and \(R_{px}\) are the horizontal positions, in pixels, of the object in the left and right images, respectively.

\begin{equation}
\vspace{-0.5em}
d = \frac{f S}{|L_{px} - R_{px}|}
\label{objdistance}
\end{equation}

Then, the horizontal displacement of the object, \(d_h\), in the direction of \(\uvec{h}\), from the center of the camera's view is calculated using equation \eqref{dh}. This is added to \(\frac{1}{2} S\) to account for the position of the right camera offset from the center of the ZED Mini. The images used in our experiments are 720x1280, so \(R_{px}-640\) represents the horizontal location of the object relative to the center of the right image.
\begin{equation}
    \vspace{-0.5em}
    d_h = S (\frac{1}{2} + \frac{(R_{px}-640)}{|L_{px}-R_{px}|})
    \label{dh}
\end{equation}

Likewise, the vertical position, \(R_{py}\), of the object in the right image is used to calculate the vertical displacement of the object, \(d_v\), in the direction of \(\uvec{v}\), from the center of the camera's view. Similarly, \(R_{py}-360\) represents the vertical location of the object relative to the center of the right image.
\begin{equation}
    \vspace{-0.5em}
    d_v = S \frac{360-R_{py}}{|L_{px}-R_{px}|}
    \label{dv}
\end{equation}

The object's position relative to the camera arm is then calculated as the position of the camera added to the distances and displacements multiplied by their respective unit vectors, as shown in equation \eqref{pobject}.
\begin{equation}
    \vspace{-0.5em}
    P_{object} = P_{camera} + d \uvec{n} + d_h \uvec{h} + d_v \uvec{v}
    \label{pobject}
\end{equation}

\subsection{Control}
\label{subsection: Control}
The Control module uses the list of the objects' positions relative to the camera arm to calculate the centroid of the objects as their average position. An image plane containing the centroid and defined by \(\uvec{v}\) and \(\uvec{h}\) is constructed. The distance from the camera to the image plane, \(d_{cam}\), is calculated using equation \eqref{dcam} where \(\vec{p}_{centroid}\) and \(\vec{p}_{camera}\) represent vectors from the origin to the centroid and camera's position, respectively. The camera's position on the image plane defines the origin of the image plane. 
\begin{equation}
    \vspace{-0.5em}
    d_{cam} = (\uvec{n} \cdot \vec{p}_{centroid}) - (\uvec{n} \cdot \vec{p}_{camera})
    \label{dcam}
\end{equation}

Then, each object is projected onto the image plane and the average distance from the centroid to each object's projected position is calculated. The height of the image plane within view of the camera is calculated using equation \eqref{vertvis}, proportional to the distance from the image plane to the camera. The radius of the desired view was set at 50\% and defined a circle that should maximally overlap the circle drawn around the centroid by the average distance to the objects.
\begin{equation}
    \vspace{-0.5em}
    h_{visable} = d_{cam} tan(30^\circ)
    \label{vertvis}
\end{equation}

Based on the location of the centroid with respect to the origin of the image plane (the center of the camera's view), the control rules determine the movement of the camera. Zoom is controlled by the difference in size of the circles around the centroid and origin. If the average distance from the centroid to the objects is larger than the desired view ring, then the system proposes zooming out. Conversely, if the average distance from the centroid to the objects is smaller than the desired view ring, then the system proposes zooming in.  The tilt and pan of the camera are controlled by the location of the centroid on the image plane relative to the origin of the image plane. If the centroid is further away from the origin than 30\% of the visible height of the camera's view in any direction, then the camera is tilted up or down, or panned left or right to bring the centroid closer to the center of the camera's view.

\section{Experimental Evaluation}
The autonomous camera system was evaluated by examining the Perception and Control modules separately as well as assessing the system as a whole. This analysis considered 27 pairs of left and right images of a block transfer task showing the left and right graspers and several blocks. These 54 images were annotated to create the ground truth of the locations of the objects in the images. The accuracy of the MRCNN was defined as the ratio of correctly localized and classified objects to the total number of objects present in the image set. An object was considered correctly localized if the center of the object was within the bounding box generated by the MRCNN. 
The Root Mean Square Error (RMS) for the horizontal and vertical locations of the centers of the objects were also calculated. To evaluate the Control module, 13 students were shown these 27 pairs of images and asked how they would adjust the camera's zoom, pan, and tilt to achieve a better view of the objects in the environment. For each pair of images, the commands given by the Control module using the object locations directly from the Perception module and the commands given by the Control module using the ground truth object locations were compared to the majority vote of the commands from the survey. 

All the experiments were conducted on an x86\_64 PC with an Intel Core i7 CPU @ 3.70GHz and 32GB RAM, running Linux Ubuntu 18.04 LTS, and an Nvidia 1080 Ti GPU, running CUDA 10.1. We used Keras \cite{chollet2015keras} API v.2.2.4 on top of TensorFlow \cite{abadi2016tensorflow} v.1.13.1 for training our model and Scikit-learn \cite{pedregosa2011scikit} v.0.21.3 for pre-processing and evaluation.

\subsection{Perception}

\begin{figure*}[htbp]
    \centering
    \subfloat[Left image from the ZED Mini]{\includegraphics[width=0.40\textwidth]{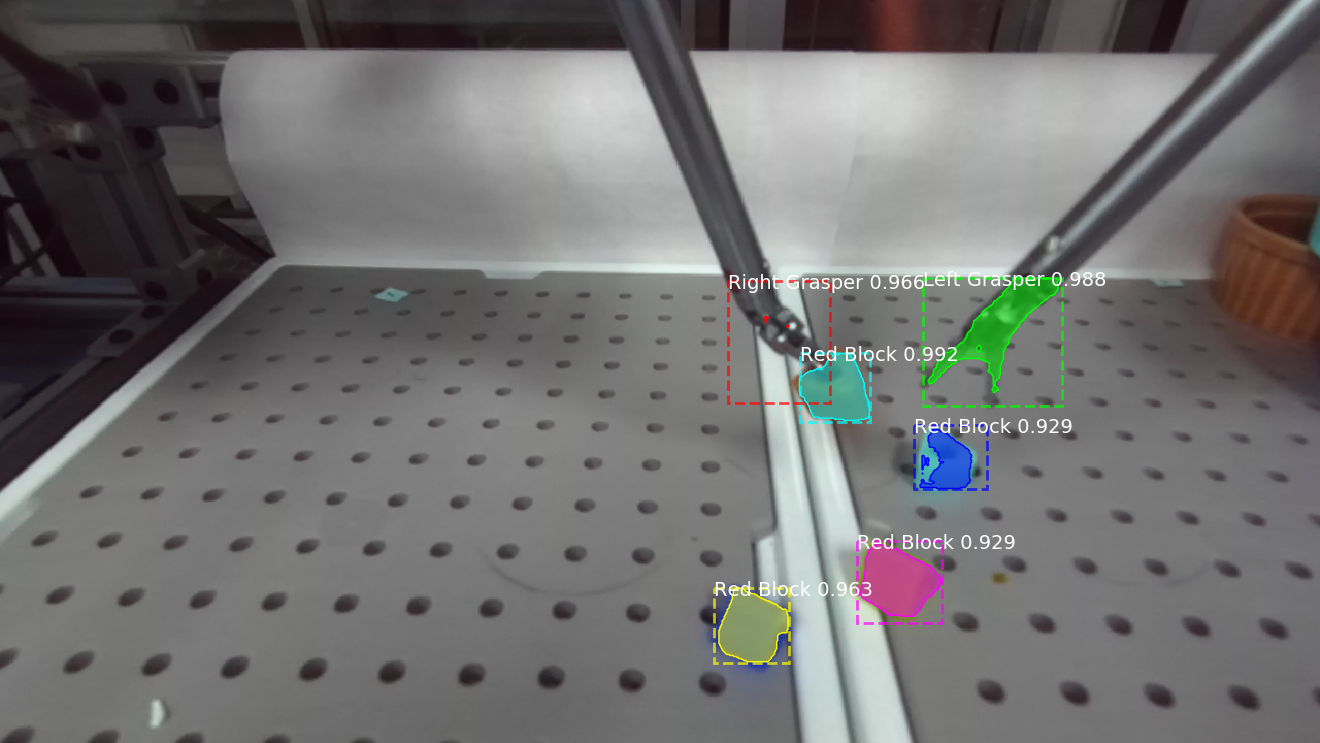}\label{fig:left}}
    \hspace{1em}
    \subfloat[Right image from the ZED Mini]{\includegraphics[width=0.40\textwidth]{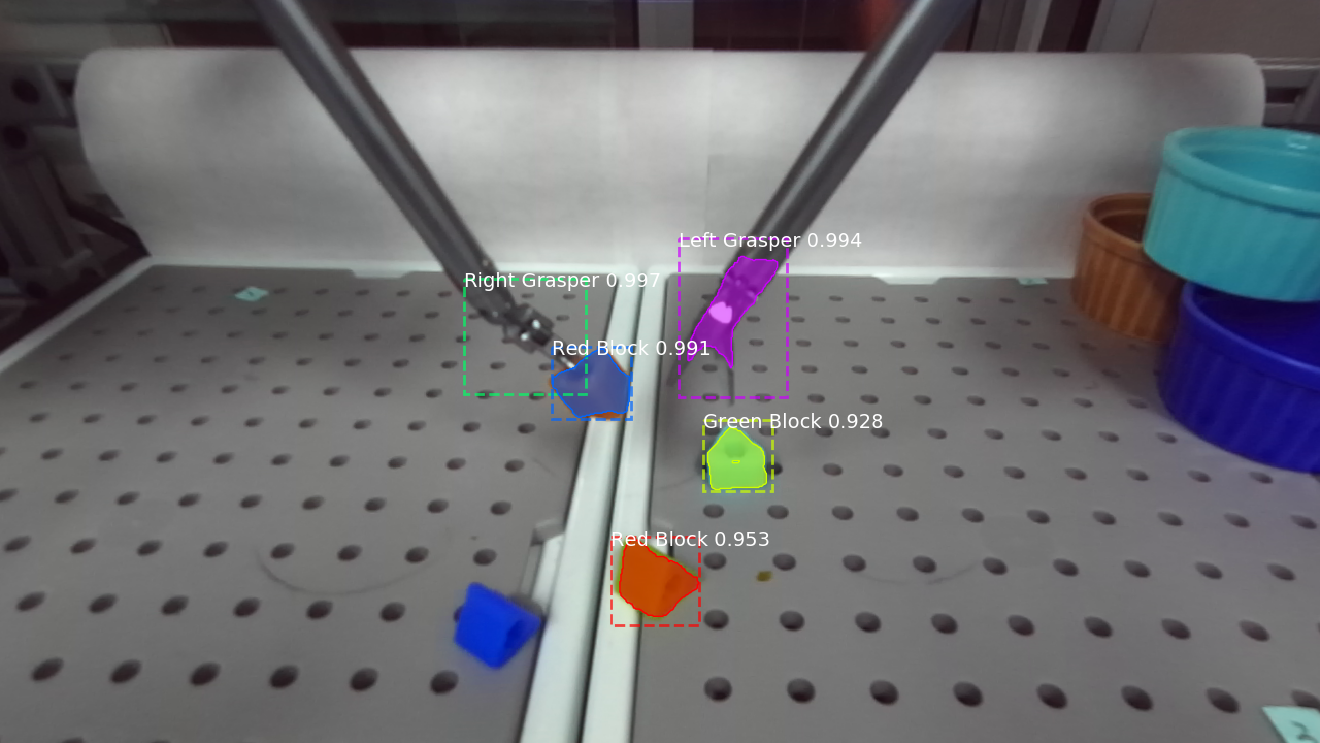}\label{fig:right}}
    \vspace{-0.5em}
    \caption{Pair of images with objects localized and classified by the MRCNN}
    \label{splash}
    \vspace{-1.5em}
\end{figure*}

The model was evaluated in terms of its overall loss, bounding box loss, classification loss, validation bounding box loss, and validation classification loss which are listed in Table \ref{MRCNNlosses}. The validation bounding box loss was 0.1346, and the validation class loss was 0.0274. 
We used hyper-parameter tuning to find the optimum learning rate which was 0.01 for 20 epochs, each with step-size of 50 and a batch size of 1 image. Tensorboard was used to visualize the impact that certain hyperparameters had on the model's performance.

\begin{table}[htbp]
\vspace{-1em}
\caption{MRCNN Losses}
\vspace{-1em}
\begin{center}
\begin{tabular}{|c|c|}
\hline
Type & Loss \\
\hline \hline
Bounding Box Loss & 0.0422 \\
\hline
Class Loss & 0.0312 \\
\hline
Validation Bounding Box Loss & 0.1346 \\
\hline
Validation Class Loss & 0.0274 \\
\hline
Overall Loss & 0.2672 \\
\hline
\end{tabular}
\label{MRCNNlosses}
\end{center}
\vspace{-2em}
\end{table}

The MRCNN model predicted on images from the ZED Mini and returned a list of objects identified for each image. One such pair of left and right images is shown in Fig.~\ref{splash}. Although both graspers and four blocks were localized in the left image, only the two graspers and three blocks were detected in the right image. The rightmost block in the right image was classified as a ``Green Block" and was ignored by the algorithm. In addition, the vertical difference in the positions of the right grasper in the left and right images was too large, so it was not correctly paired, and was thus also ignored. This left us with only three objects to use in the proceeding calculations, the left grasper and two blocks.

The MRCNN consistently misclassified the left grasper as the ``Right Grasper" and the right grasper as the ``Left Grasper". Errors in the classification of the graspers can be attributed to insignificant differences between the color of the graspers and the background. The graspers appear to blend in with their background and although apparent to the naked eye, the two can be easily confused with a camera. The right and left graspers were also very similar in shape and size, so without information about their relative position in regards to other objects, they would be difficult to differentiate.

Different colored blocks were also used to test the ability of the MRCNN to discern color, but since most of the blocks were classified as ``Red Blocks" regardless of their color, this analysis considered ``Blocks" in general. 
The model had an accuracy of 52.87\% in correctly identifying blocks in the images. Due to the consistent mislabeling of the graspers, this analysis considered ``Graspers" in general as well. The model had an accuracy of 80.77\% in correctly identifying graspers in the images. Table \ref{objclassacc} shows the number of objects correctly classified out of the total number of objects for all 54 images. The MRCNN correctly localized but misclassified seven objects and in two cases incorrectly identified the frame of the Raven II as a ``Right Grasper". Overall, the model had an accuracy of 61.21\% in correctly classifying blocks and graspers in the images and provided sufficient information for the Control module to estimate the centroid of the objects and propose commands to move the camera.

\begin{table}[htbp]
\caption{Object Classification Accuracies}
\vspace{-1.5em}
\begin{center}
\begin{tabular}{|c|c|c|c|}
\hline
Class & Correctly Classified & Total & Accuracy (\%) \\
\hline \hline
Graspers & 84 & 104 & 80.77\\
\hline
Blocks & 129 & 244 & 52.87\\
\hline
All objects & 213 & 348 & 61.21 \\
\hline
\end{tabular}
\label{objclassacc}
\end{center}
\vspace{-1.5em}
\end{table}

The ground truth annotations were also used to determine the RMS in the horizontal and vertical locations of the centers of the objects in each image. The RMS error in the horizontal and vertical locations of the bounding boxes proposed by the MRCNN and the centers of the ground truth annotations were 20.37 pixels and 16.98 pixels, respectively. The sorting and pairing algorithm used by the system to match the position of an object in the left image with its position in the right image tolerated up to a 20 pixel difference in vertical position, so this was an acceptable amount of error. 

\subsection{Control}
\begin{figure*}[t!]
    \centering
    \subfloat[]{\includegraphics[width=0.35\textwidth]{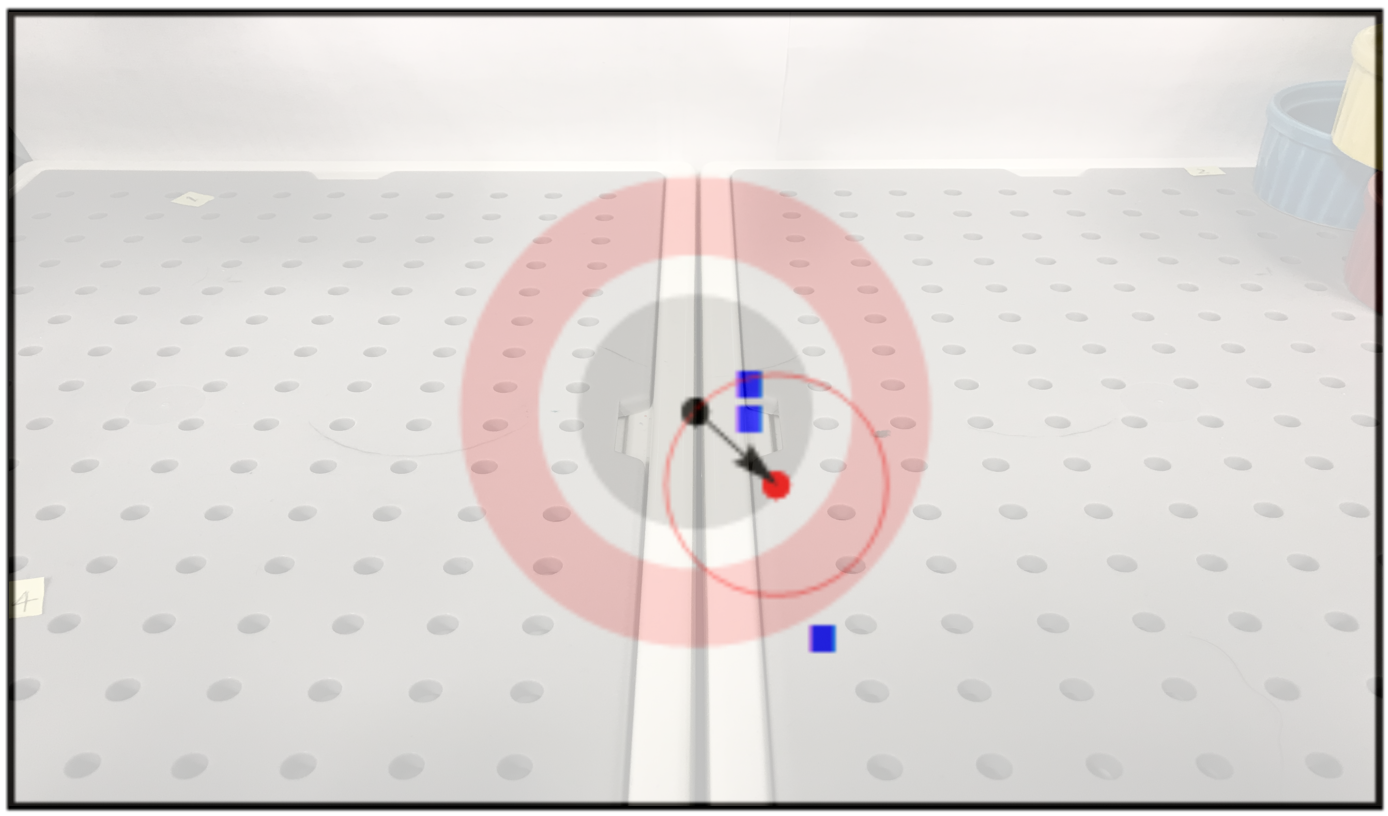}\label{fov3}}
    \hfill
    \subfloat[]{\includegraphics[width=0.35\textwidth]{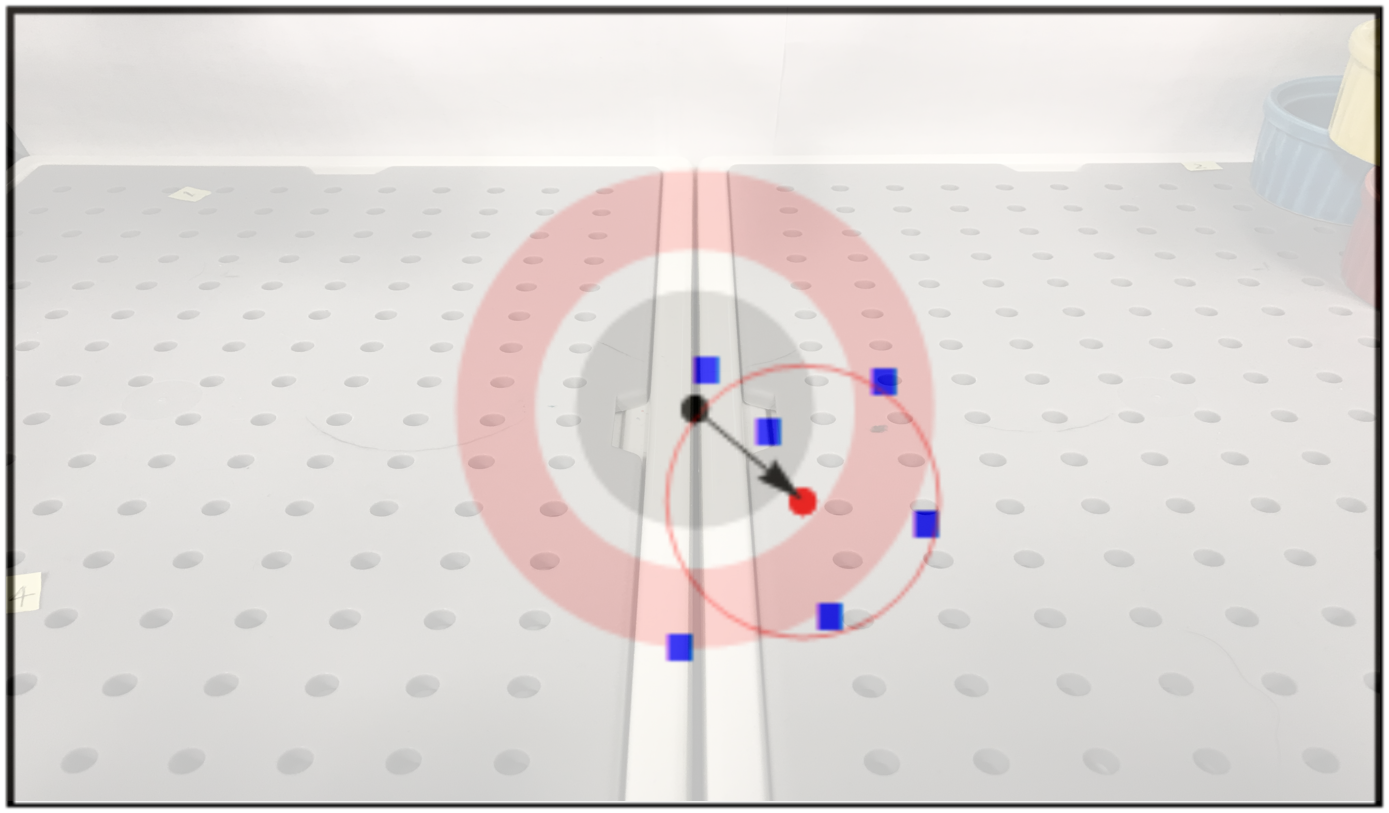}\label{fovgt3}}
    \hfill
    \subfloat{\includegraphics[width=0.17\textwidth]{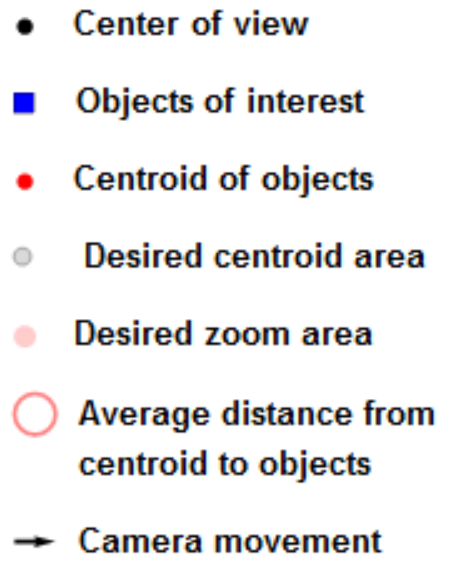}}
    \vspace{-1em}
    \caption{Projection of objects located in Fig.~\ref{splash} onto the camera image plane. (a) Object coordinates directly from MRCNN where mispairing resulted in incorrect position estimations, (b) Object coordinates from ground truth.}
    \label{imageplane}
    \vspace{-1em}
\end{figure*}

Fig.~\ref{imageplane} shows the objects in Fig.~\ref{splash} mapped to the image plane. Given the estimated positions of the objects identified by the Perception module, the Control module mapped them to the image plane as shown in Fig.~\ref{fov3}. The Control module sent commands to the IBIS to zoom in.
However, given the list of estimated positions for each object calculated using the ground truth annotations, the Control module mapped these objects to the image plane as shown in Fig.~\ref{fovgt3} and proposed zooming in, tilting down, and panning right instead. 

\begin{figure}[b!]
    \vspace{-3em}
    \centering
    \subfloat[]{\includegraphics[width=0.2\textwidth]{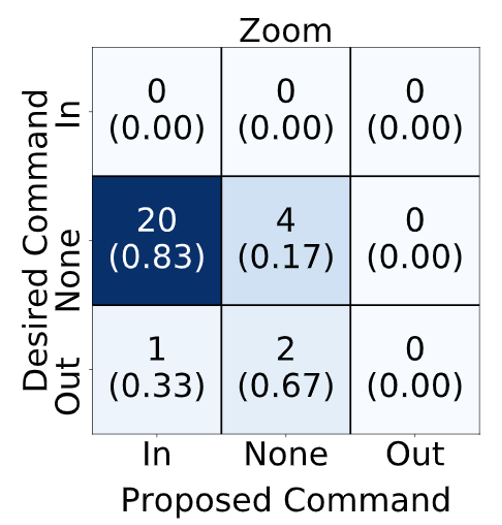}\label{zoom}}
    \subfloat[]{\includegraphics[width=0.2\textwidth]{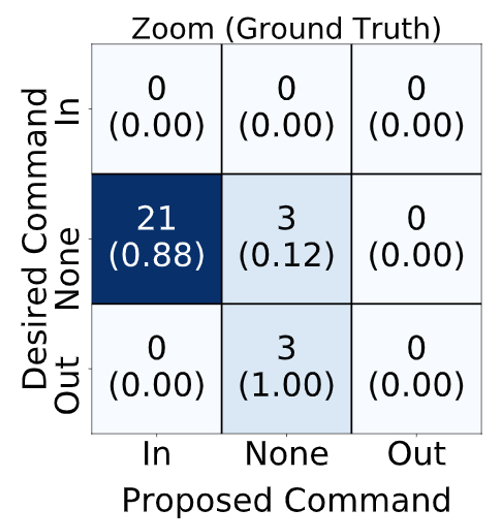}\label{zoomgt}}
    \vspace{-1em}
    \hfill
    \subfloat[]{\includegraphics[width=0.2\textwidth]{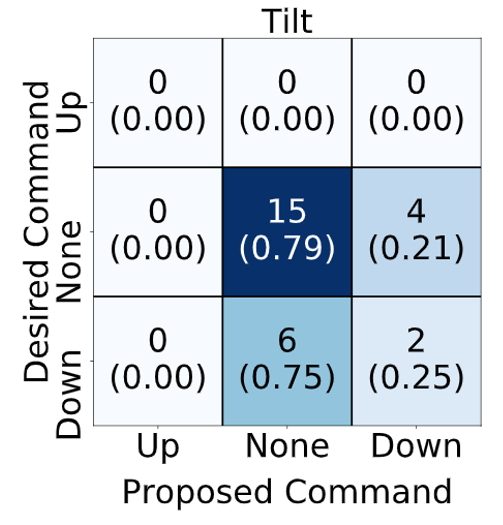}\label{tilt}}
    \subfloat[]{\includegraphics[width=0.2\textwidth]{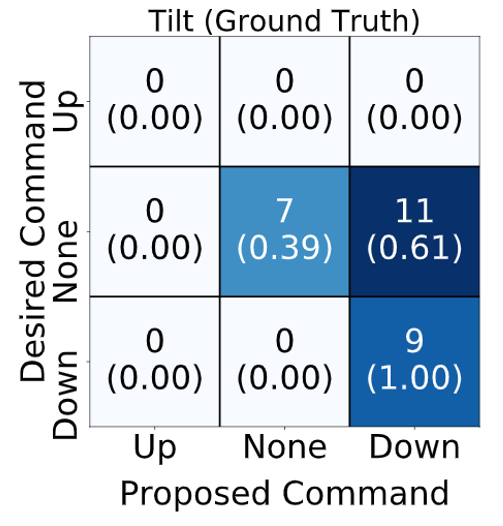}\label{tiltgt}}
    \vspace{-1em}
    \hfill
    \subfloat[]{\includegraphics[width=0.2\textwidth]{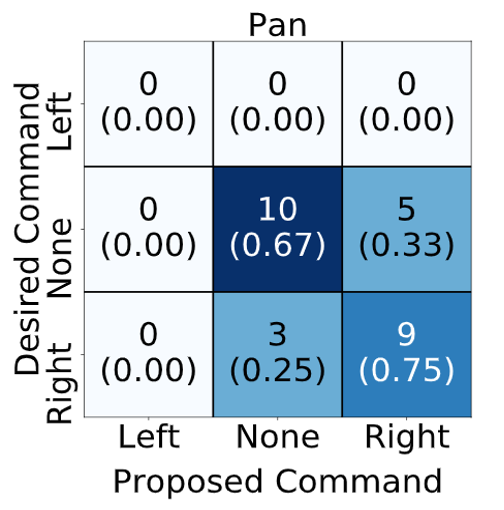}\label{pan}}
    \subfloat[]{\includegraphics[width=0.2\textwidth]{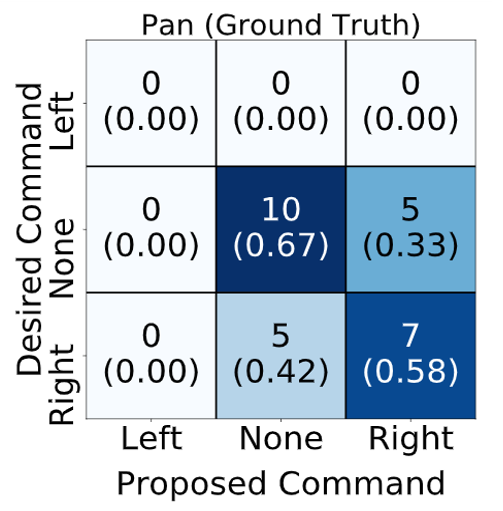}\label{pangt}}
    \vspace{-1em}
    \caption{Confusion matrices for Control module commands. (a), (c), and (e) are based on MRCNN predictions. (b), (d), and (f) are based on ground truth annotations for Perception.}
    \label{confusionmatrices}
    \vspace{-1em}
\end{figure}

The Control module determined commands to move the camera based on the location of the centroid relative to the center of the camera's view, and the average distance between the centroid and each object compared to a desired zoom radius. In Fig.~\ref{imageplane}, the center of the camera's view is a black dot while the centroid of the objects is a red dot. The gray disk represents the desired area that the centroid should be in and the Control module adjusted the pan and tilt of the camera to move the gray disk towards the centroid. The average distance between the centroid and objects is shown with a red circle around the centroid and represents the size of the region of interest. The thick pink ring around the center of the image represents the desired zoom area 
and the Control module adjusted zoom so that the red circle was within the pink ring. 

In order to evaluate the control rules used to determine camera movement commands, 13 graduate and undergraduate students were shown the 27 pairs of images and asked to select commands for zoom, tilt, and pan (no movement was also an option for each category). The ground truth for desired movements for each image was determined by majority voting based on the survey responses.
The confusion matrices between the desired commands and the Control module's commands are shown in Fig.~\ref{confusionmatrices}. Figures \ref{zoom}, \ref{tilt}, and \ref{pan} show the confusion matrices for the commands proposed by the system when directly fed the predictions of the MRCNN, and Figures \ref{zoomgt}, \ref{tiltgt}, and \ref{pangt} show the confusion matrices for the commands proposed by the system based on the ground truth annotations for the Perception module (assuming perfect Perception).

The set of images used in this analysis included frames showing a block transfer task, during which the camera position was held constant. The task occurred in the central, lower, and slightly right regions of the operating space, which meant appropriate camera commands were limited to tilting down, panning right, adjusting zoom, or no movement. Thus, the confusion matrices in Fig.~\ref{confusionmatrices} only show data for these movements. The confusion matrices for the zoom command show that the system proposed zooming in when the desired command was no movement. This behavior was consistent even when the Control module was given the ground truth annotations. Since the amount of zoom tends to be subjective, the desired zoom area should be adjustable to accommodate personal preferences. The confusion matrices for the tilt command show that in 75\% of cases the system proposed no adjustments to tilt even if the desired command was to tilt down. However, when given the ground truth annotations, the system often proposed tilting down when the desired command was no movement (61\%). On the other hand, the system usually selected appropriate commands for panning, even given the ground truth annotations. 

The undesired tilt down commands could be explained by the equal weighting of all objects in the centroid calculation. The Perception module had a higher accuracy in detecting graspers, which were usually located more centrally in the images and would have led the system to propose no change in tilt. But using the ground truth, the blocks would have pulled the location of the centroid down in the image resulting in tilt down commands. The difference between Fig.~\ref{tilt} and \ref{tiltgt} suggests that a weight function should be implemented in the centroid calculation to address how some objects are more important to view than others.

\section{Conclusion}
This work presents a proof of concept of an autonomous camera system for tele-operated robotic surgery that tracks the centroid of all objects of interest in the field of view of the camera. The objects of interest were identified using transfer learning, with an MRCNN partly pre-trained on the COCO dataset. A custom-built camera arm was created for positioning the camera and the centroid of objects of interest was tracked using a set of control rules. The system was evaluated using a dataset of images from dry-lab experiments and by comparing the proposed motions of the camera to the desired motions. 
The evaluation results suggest that the system proposes appropriate movements for the tilt and pan of the camera, but the desired zoom area should be decreased to provide a wider field of view and an object weight function should be implemented in the centroid calculation. Future work will focus on improving the accuracy of the Perception module by increasing the size and diversity of the training set and generating more accurate annotations, adjusting the control rules for zoom, and testing the system on a wider variety of object configurations and movement directions.


\section*{Code Availability}
Designs and code for the IBIS and the perception model are respectively available at \url{https://github.com/kch4fk/IBIS} and \url{https://github.com/thatssoharsh/Raven-II-Perception}.

\section*{Acknowledgment}
We would like to thank Erick Tian, Asha Maran, and Ray Wang for their contributions to the collection and analysis of data from the IBIS camera arm. 
This work was partially supported by an award from the Graduate Medical Education (GME) Innovation grant program at the University of Virginia Health System and the National Science Foundation NRT program under Grant No. 1829004. 

\vspace{12pt}
\bibliographystyle{IEEEtran}
\bibliography{references.bib}
\end{document}